\documentclass{article}
\usepackage{spconf,amsmath,graphicx}
\usepackage{xcolor}
\usepackage{soul}

\usepackage{subfig}

\title{Improving Speech Recognition on Noisy Speech via Speech Enhancement with Multi-Discriminators CycleGAN}
\name{Chia-Yu Li and Ngoc Thang Vu}
\address{Institute of Natural Language Processing, University of Stuttgart, Germany}
\begin{document}
%
\maketitle
\begin{abstract}
This paper presents our latest investigations on improving automatic speech recognition for noisy speech via speech enhancement. We propose a novel method named Multi-discriminators CycleGAN to reduce noise of input speech and therefore improve the automatic speech recognition performance. Our proposed method leverages the CycleGAN framework for speech enhancement without any parallel data and improve it by introducing multiple discriminators that check different frequency areas. Furthermore, we show that training multiple generators on homogeneous subset of the training data is better than training one generator on all the training data. We evaluate our method on CHiME-3 data set and observe up to 10.03 \% relatively WER improvement on the development set and up to 14.09 \% on the evaluation set.
\end{abstract}
\section{Introduction}
Single-channel speech enhancement (SE) strives for reducing the noise component from noisy speech to increase the intelligibility and perceived quality of the speech component \cite{SE}. It has been widely used as a pre-processing in speech-related applications such as automatic speech recognition (ASR). In the past years, deep learning has been employed in single-channel SE and has achieved great success. Some works proposed deep learning based mask to filter out the noise on noisy input features \cite{deepMask1,deepMask2,deepMask3}.  However, the mask approach is based on unrealistic presumption: the noise is strictly additive and the scale of the masked signal is the same as the clean target. Therefore, the feature mapping approach was introduced to deal with that problem by training a mapping network that direct transforms the noisy features to the clean ones \cite{FM1,FM2,FM3,FM4,FM5,FM6}.  

Generative Adversarial Network (GAN) \cite{GAN}, Adversarial Training \cite{DAT} and Cycle-Consistent Adversarial Networks (CycleGAN) \cite{CycleGAN2017} have drawn attention in the deep learning community since they demonstrate better generalization by using discriminator network to encourage the model to produce noise-invariant features in order to ease the mismatch issue between source and target domain. These approaches have also applied to SE \cite{CFL-SE-DAT,ZM-DAT,ZM-CycleGANSE}. Meng et al \cite{ZM-CycleGANSE} proposed cycle-consistent speech enhancement (CSE) and adversarial cycle-consistent speech enhancement (ACSE) models in which an additional inverse mapping network is introduced to reconstruct the noisy features from the enhanced ones. Furthermore, a cycle-consistent constraint is enforced to minimize the reconstruction loss. CSE and ACSE models are designed for both, training with parallel and non-parallel data. The main difference between CSE and ACSE architecture is that ACSE used two discriminator networks to distinguish the enhanced and noised features from the noisy and clean features \cite{ZM-CycleGANSE}. Evaluated on the CHiME-3 dataset, the CSE model achieves reasonable relative word error rate (WER) improvement, while the ACSE model, which is trained with unparalleled data, is less effective. 

Generative multi-adversarial network (GAM), a framework that extends GAN to multiple discriminators, was proposed in Computer Vision \cite{GMAN}.  GMAN seems to produce higher quality samples in a fraction of the iterations when measured by a pairwise GAM-type metric \cite{GMAN}. Hosseini-Asl et al \cite{HS-Multi-D} proposed a Multi-Discriminators CycleGAN (MD-CycleGAN), similar to generative multi-adversarial network (GMAN), for unsupervised non-parallel speech domain adaptation. The MD-CycleGAN model employs multiple independent discriminators on the power spectrogram, each in charge of different frequency bands \cite{HS-Multi-D}. It demonstrates the effectiveness of MD-CycleGAN method on CTC End-to-End ASR with gender adaptation \cite{DBLP:journals/corr/abs-1712-07101}. However, the input and the generated features are the power spectrogram, which is not common for the state-of-the-art ASR system (hyrid or E2E). Besides, MD-CycleGAN did not take into account the identity loss which is meaningful in non-stationary noise scenario, which we will explain it in more detail in section 2.

In this paper, we contribute to previous work in the following aspects: 1) To the best of our knowledge, we are the first to propose a novel framework based on the Multi-Discriminators generative models \cite{GMAN,HS-Multi-D} for feature mapping in speech enhancement and the feature mapping model uses the same input as ASR (log-Mel filterbank) ; 2) We propose to train multiple generators and multiple discriminators on homogeneous data to improve the WER  3) We show that our models outperform strong baselines with up to 10.03 \% and up to 14.9 \% relatively WER improvements on CHiME-3 development set and evaluation set without retraining the ASR system that was trained with WSJ clean data. 

\section{Method}
\subsection{CycleGAN}
The goal of CycleGAN is to solve the image to image translation task when learning a mapping G\_A between an input image from a source domain A and an output image from a target domain B without using paired training data \cite{CycleGAN2017}. The mapping $G\_A: A \rightarrow B$ is learnt such that the distribution of images from $G\_A(A)$ is indistinguishable from the distribution $B$ using an adversarial loss. Because this mapping is highly under-constrained, \cite{CycleGAN2017} coupled it with an inverse mapping $G\_B: B \rightarrow A$ and introduce a cycle consistency loss to push $G\_B(G\_A(A))=A$ (and vice versa). The full objective is as follows:
\begin{align}
    L = L_{G\_A} + L_{G\_B}
    + \lambda_{idt} * L_{idt\_A} + \lambda_{idt} * L_{idt\_B}\nonumber\\
    + \lambda_{cycle} * L_{cycle\_A}+ \lambda_{cycle} * L_{cycle\_B}\label{eq:eq1}
\end{align}
where $\lambda_{idt}$ and $\lambda_{cycle}$ are tunable parameters. And other losses are defined as follows:
\begin{align}
   L_{G\_A} = MSELoss(D\_A(G\_A(A)), True)\label{eq:eq2}\\
   L_{G\_B} = MSELoss(D\_B(G\_B(B)), True)\label{eq:eq3}\\
   L_{idt\_A} = L1Loss(G\_A(B), B)\label{eq:eq4}\\
   L_{idt\_B} = L1Loss(G\_B(A), A)\label{eq:eq5}\\
   L_{cycle\_A} = L1Loss(G\_B(G\_A(A)),A)\label{eq:eq6}\\
   L_{cycle\_B} = L1Loss(G\_A(G\_B(B)),B)\label{eq:eq7}
\end{align}
The loss functions for two discriminators $D\_A$ and $D\_B$ are
{\scriptsize
\begin{align}
    L_{D\_A} = \frac{MSELoss(real\_B, True) + MSELoss(fake\_B, False)}{2}\label{eq:eq8}\\
    L_{D\_B} = \frac{MSELoss(real\_A, True) + MSELoss(fake\_A, False)}{2}\label{eq:eq9}
\end{align}
}
Note that the outputs of $G\_A(A)$ and $G\_B(B)$ are also called \textit{fake\_B} and \textit{fake\_A},respectively. Compared with the objective of ACSE model \cite{ZM-CycleGANSE}, the main difference is that ACSE does not take into account the identity loss as shown in equations (\ref{eq:eq4}) and (\ref{eq:eq5}). However, the noise does not constantly occur in every frame in the non-stationary noise scenario. Therefore, adding identity loss to the objective could benefit the model to generate original frame when the input feature is in the output domain. That is, $G\_A(B)=B$ and $G\_B(A)=A$.

\begin{figure}[!htb]
    \begin{center}
        \includegraphics[scale=0.25]{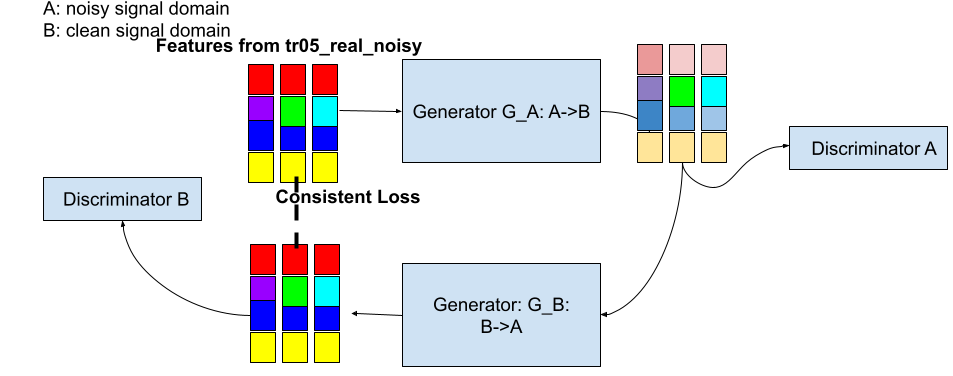}
    \end{center}
    \caption{The architecture of CycleGAN }
    \label{fig:cyclegan}
\end{figure}

Figure \ref{fig:cyclegan} shows the network architecture of CycleGAN for converting a noisy signal to a clean signal. $A$ and $B$ are the domain of noisy signal and clean signal, respectively. The two generators are $G\_A: A \rightarrow B $ and $G\_B: B \rightarrow A$. The goal of the two discriminators ($D\_A$ and $D\_B$) is to predict whether the sample is from the actual distribution ( `real') or produced by the generator (`fake') given the feature input . 

\subsection{Multi-Discriminators CycleGAN}
\begin{figure}[!htb]
    \begin{center}
        \includegraphics[scale=0.25]{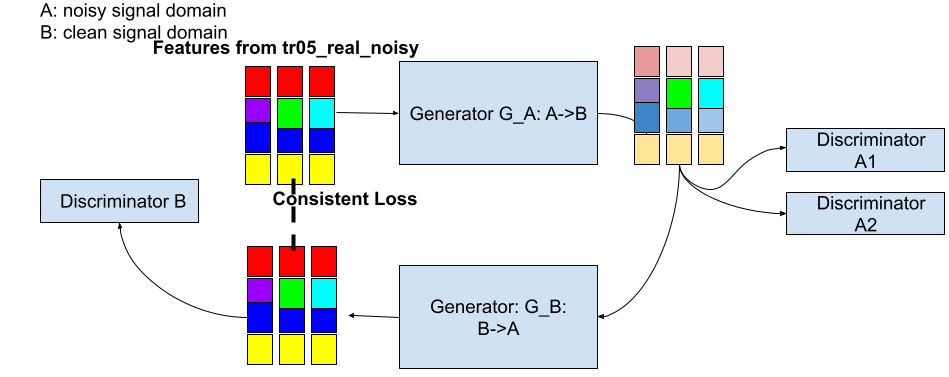}
    \end{center}
    \caption{The architecture of Multi-Discriminators CycleGAN (e.g. n=2) }
    \label{fig:cycle-2da}
\end{figure}

The model is based on Multi-Discriminators generative models \cite{GMAN,HS-Multi-D} and CycleGAN \cite{CycleGAN2017}. 
Based on equation (\ref{eq:eq2}), the loss function for generator $G\_A$ is formed of the prediction of discriminator $D\_A$. That is to say, if the discriminator is imperfect, then it can not guide the generator to the optimal area during the training. Therefore, we propose to strengthen the complexity of the discriminator by introducing multiple discriminators, e.g. $D\_A_{1}$ and $D\_A_{2}$, to judge the $G\_A(A)$ (a.k.a fake\_B) as shown in Figure \ref{fig:cycle-2da}. Each discriminator is responsible for a subregion of the log Mel-filterbank feature. The full objective is the same as equation \eqref{eq:eq1}, but the equation \eqref{eq:eq2} is adapted to 
\begin{align}
    L_{G\_A} = \frac{\sum_{i=1}^{n} MSELoss(D\_A_{i}(G(A)), True)}{n}\label{eq:eq10}
\end{align} 
Where $n$ is the total number of discriminators to judge $G\_A$ output. The loss function for each discriminator in equation (\ref{eq:eq8}) is adapted to
{\footnotesize
\begin{align}
    L_{D\_A_{i}} &= (MSELoss(mask\_i(real\_B), True) \nonumber\\&+ MSELoss(mask\_i(fake\_B), False))/2\label{eq:eq11}
\end{align}
}
The $mask\_i$ is to mark out some bins in log Mel-filterbank feature, e.g., the mask for $D\_A_{1}$ marks off the 21th to 40th bins and the mask for $D\_A_{2}$ marks off the 0 to 20th bins.
\begin{align}
    step=feat\_dim/n\\
    start=(i-1)*step\\
    end=i*step\\
    mask\_i=input[start:end]
\end{align}
Furthermore, we can even increase the number of generators and therefore the number of discriminators by dividing the training data in different homogeneous subsets, e.g. by splitting the data by genders and by types of noises.

\section{DataSet}
CHiME-3 was developed as part of the 3rd CHiME Speech Separation and Recognition Challenge. It contains approximately 342 hours of English speech and transcripts from noisy environments and 50 hours of noisy environment audio. The CHiME Challenges focus on distant-microphone ASR in real-world environments \cite{CHiME3}. Table \ref{tab:chime3} shows that the training set contains 8738 noisy utterances: 1600 real noisy utterances from two male and two female speakers in the 4 noisy environments and 7138 simulated noisy utterances based on WSJ0 SI-84 training set in the 4 noisy environments \cite{CHiME3}. Development set is composed of 410 (real noisy speech) $\times$ 4 (environments) + 410 (simulated noisy speech) $\times$ 4 (environments) = 3280 utterances. Test set consists 330 (real noisy speech) $\times$ 4 (environments) + 330 (simulated noisy speech) X 4 (environments) = 2640 utterances \cite{CHiME3}.
\begin{table}[]
    \centering
    \footnotesize
    \begin{tabular}{|l|c|c|c|c}
         \hline
          Train set  & Noisy type & \# of utterance & hr. \\\hline
         tr05\_orig\_clean  &  N/A & 7138 & 15 \\\hline
         tr05\_real\_noisy &  BUS,CAF,PED,STR & 1600 & 2 \\\hline
         tr05\_simu\_noisy &  BUS,CAF,PED,STR & 7138 & 15 \\\hline\hline
         Test set & Noisy type & \# of utterance & hr. \\\hline
         dt05\_real\_noisy & BUS,CAF,PED,STR & 1640 & 2.74 \\\hline
         dt05\_simu\_noisy & BUS,CAF,PED,STR & 1640 & 2.89 \\\hline
         et05\_real\_noisy & BUS,CAF,PED,STR & 1640 & 2.17 \\\hline
         et05\_simu\_noisy & BUS,CAF,PED,STR & 1640 & 2.27 \\\hline
    \end{tabular}
    \caption{CHiME-3 dataset}
    \label{tab:chime3}
\end{table}
\section{Experimental Setup}
\subsection{Baseline ASR}
The ASR is trained with 80 hours of Wall Street Journal \cite{wsj} using Kaldi \cite{Povey_ASRU2011} default recipe to train HMM-GMM model to get alignment, and then we train the time delay neural network (TDNN) network \cite{tdnnkaldi} with the augmented features (speed and volume perturbed). The input feature consists of 40-bin log-Mel filterbank and 100-bin ivectors. In decoding stage, we use a fairly large dictionary, a trigram pruned language model and run re-scoring with fourgram language model. 

\subsection{Multi-Discriminators CycleGAN}
This section mainly describes the network structure of Multi-Discriminators CycleGAN, the experimental setup and how we train it with CHiME-3 dataset. The experimental setup and code could be obtained in github\footnote{https://github.com/chiayuli/SEWork}.

The generators and discriminators are trained with 3200 hours of data containing real noisy train set (tr05\_real\_noisy) and clean data (tr05\_orig\_clean). The feature is 40-bin log-Mel filterbank using hamming window. The context window of each frame is 5, so the input size for the generator and the discriminator is (1, 11, 40). The network for the generator is 9 blocks Resnet \cite{DBLP:journals/corr/HeZRS15} and the number of filters in the last convolutional layer in Resnet is 64. The network of the discriminator contains 3 convolutional layers with normalized layer. The number of filters in the first convolutional layer is 64. The batch size is 512 and the learning rate is 0.0002 with learning decay every 50 epochs. The optimizer is Adam \cite{adam} and the model is trained for 200 epochs. The $\lambda_{idt}$ and $\lambda_{cycle}$ are 0.5 and 10 in all the experiments.

\begin{figure}[!htb]
    \centering
    \includegraphics[scale=0.33]{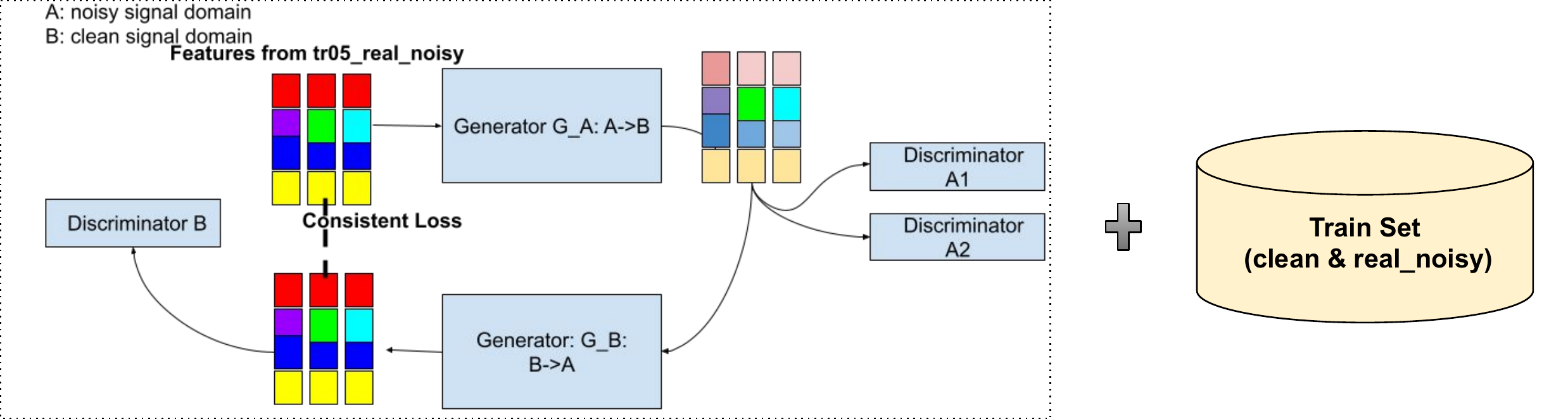}
    \caption{A1: The architecture of CycleGAN-1G+2DA}
    \label{fig:1g-model}
\end{figure}

We design three architectures in this experiment. In the first architecture (A1), we train one single generator using the Multi-Discriminators CycleGAN with all the training set as shown in Figure \ref{fig:1g-model}, i.e. 3200 hours of unpaired speech data coming from all genders and all different types of noises. For this generator, we train up to three discriminators, e.g. CycleGAN-1G+2DA means one generator and two discriminators. 

\begin{figure}[!htb]
    \centering
    \includegraphics[scale=0.33]{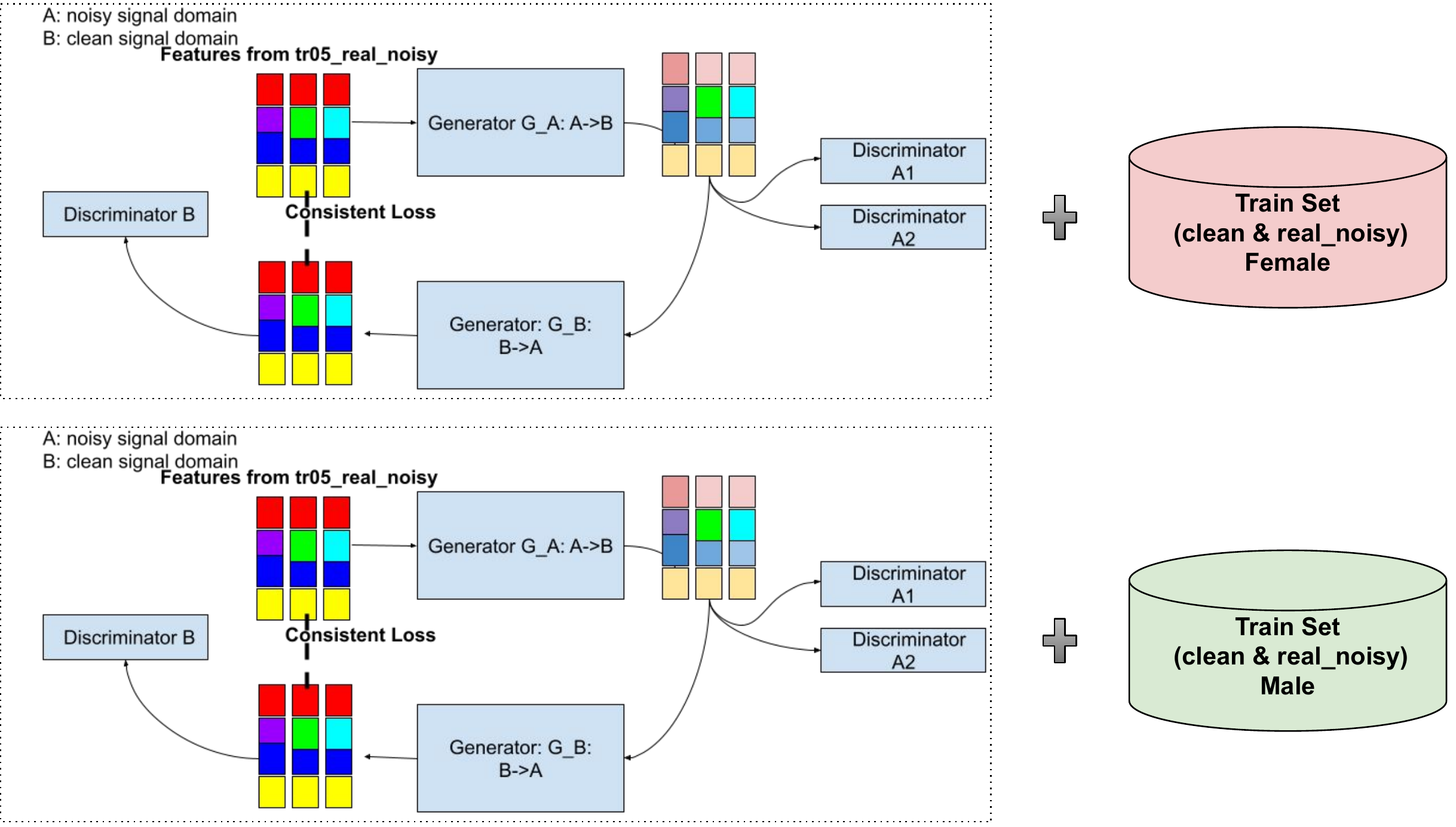}
    \caption{A2: The architecture of CycleGAN-2G+4DA}
    \label{fig:2g-model}
\end{figure}

In the second architecture (A2), we split the data by gender and use two Multi-Discriminators CycleGANs to train with each of the subset simultaneously as shown in Figure \ref{fig:2g-model}. The name of architecture with prefix "CycleGAN-2G" means that there are two CycleGANs. CycleGAN-2G+2DA means that there are two discriminators ($D\_A$), one $D\_A$ for one $G\_A$. CycleGAN-2G+4DA means that there are four discriminators ($DA$), two $DAs$ for one $G\_A$. 

\begin{figure}[!htb]
    \centering
    \includegraphics[scale=0.33]{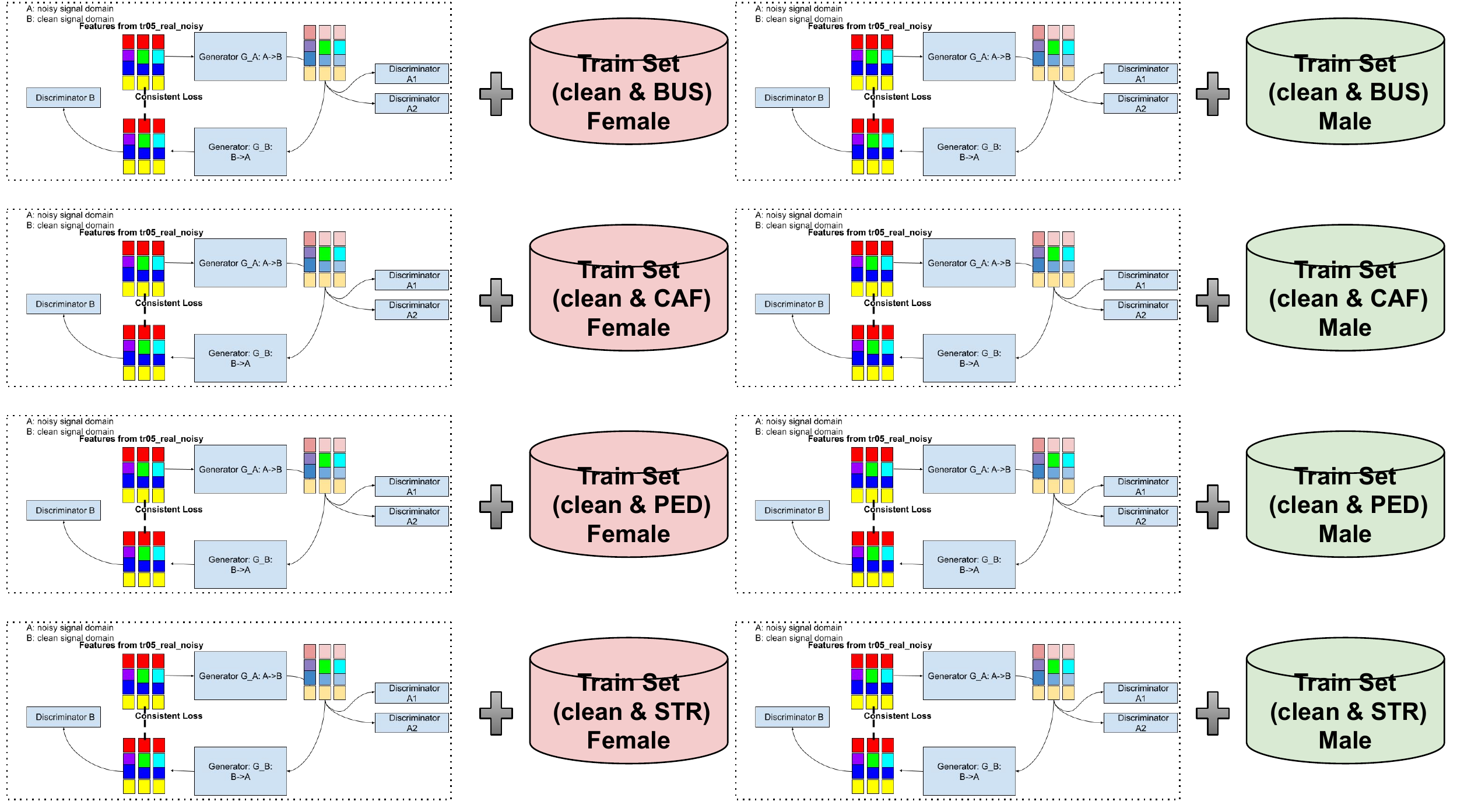}
    \caption{A3: The architecture of CycleGAN-8G+16DA}
    \label{fig:3g-model}
\end{figure}

In the last architecture (A3), we split the data not only by genders but also by types of noises. By doing so, we obtain eight subsets of training data: female \& BUS, female \& CAF, female \& PED, female \& STR, male \& BUS, male \& CAF, male \& PED, male \& STR as shown in Figure \ref{fig:3g-model}. 
We train each Multi-Discriminators CycleGAN with one of subsets simultaneously. The prefix "CycleGAN-8G" means that there are eight Multi-Discriminators CycleGANs in the structure. Hence, CycleGAN-8G+8DA means that there are eight discriminators A ($DA$), one $D\_A$ for one $G\_A$, in the structure. CycleGAN-8G+24DA means that there are 24 discriminators ($D\_A$), three $DA$ for one $G\_A$, in the structure. 

\section{Result}
Table \ref{tab:tab2} and \ref{tab:tab3} show the WERs of the ASR system in average and also broken down in different types of noises on the CHiME-3 development set and evaluation set. Note that the ASR system is trained with WSJ clean data and stays unchanged while the input features are enhanced with our proposed Multi-Discriminators CycleGAN. In both tables, the highlighted WERs are the WERs that are better than the ones without SE. The WERs in bold are the best WERs among the models with the same number of generators.
\subsection{One Discriminator vs. Multi-Discriminators}
On the development and evaluation set, the model with two discriminators for one generator A ($G\_A$) performs better than the one having only one discriminator. The model with three discriminators performs better than the one having one or two discriminators. The same observation holds for the A2:CycleGAN-2G and A3:CycleGAN-8G models. These results suggest the importance of having multiple discriminators.

However for both A1:CycleGAN-1G and A2:CycleGAN-2G models, the WERs are not better than the WERs of the baseline (no SE) in all types of noises. The WERs are only better on the BUS and the STR noisy data but not on the CAF and the PED noisy data. The best model in the architecture of A2:CycleGAN-2G on the development set is A2:CycleGAN-2G+4DA with a WER of 31.44\%. The best A2:CycleGAN-2G on the evaluation set is A2:CycleGAN-2G+6DA which achieves 57.53 WER. But we observe the best WERs with the A3:CycleGAN-8G+24DA models, i.e. 8 generators and three discriminators for each. It has a WER of 28.66\% and 52.80\% on the development set and the evaluation set, respectively and outperforms the baseline (no SE) in all types of noisy data. These results suggest the importance of having multiple generators.

\begin{table}[!htb]
\scriptsize
    \begin{tabular}{|c|l|ccccc|}
        \hline
        Arch.& Method &  AVG & BUS & CAF & PED & STR\\ \hline
        &No SE &  32.14 & 38.88 & 37.79 & 18.93 & 32.94\\ \hline
        A1&CycleGAN-1G+1DA &35.25&\hl{37.82}&43.79&22.23&\hl{31.97}\\ \hline
        A1&CycleGAN-1G+2DA &33.53&\hl{35.39}&39.47&22.84&\hl{31.03}\\ \hline
        A1&CycleGAN-1G+3DA & \textbf{\hl{29.45}} & \textbf{\hl{34.12}} &	\textbf{\hl{33.94}}&	\textbf{\hl{17.92}}&	\textbf{\hl{28.45}} \\ \hline \hline
        A2&CycleGAN-2G+2DA & 34.05 &\hl{37.11} &45.22 &23.4 &\hl{30.47}\\ \hline
        A2&CycleGAN-2G+4DA & \textbf{\hl{31.44}}& \textbf{\hl{36.14}} & 39.2 &20.22 &\textbf{\hl{30.17}}\\ \hline
        A2&CycleGAN-2G+6DA & 32.64 & \hl{37.17} & 39.57 & 21.76 & \hl{32.06}\\ \hline \hline
        A3&CycleGAN-8G-8DA & 35.12 & \hl{38.16} & \hl{37.17} & 23.06 & 38.06\\ \hline
        A3&CycleGAN-8G+16DA & \hl{30.19} &\hl{33.26} & \hl{35.91} & 19.68 & \hl{30.48} \\ \hline
        A3&CycleGAN-8G+24DA & \textbf{\hl{28.66}}& \textbf{\hl{32.00}} &\textbf{\hl{32.54}}& \textbf{\hl{18.04}} & \textbf{\hl{28.52}}\\ \hline
        A3&CycleGAN-8G+32DA & \hl{29.51}&\hl{33.6}&\hl{33.44}&\hl{18.69}&\hl{28.79} \\ \hline
    \end{tabular}
    \caption{The WERs of baseline ASR w/o and w/ Multi-Discriminators CycleGAN SE on dt05\_real\_noisy.}
    \label{tab:tab2}
\end{table}

\begin{table}[!htb]
\scriptsize
    \begin{tabular}{|c|l|ccccc|}
        \hline
        Arch.& Method &  AVG & BUS & CAF & PED & STR\\ \hline
        &No SE &  61.46 &82.91 & 62.33 & 57.21 &43.41\\ \hline
        A1&CycleGAN-1G+1DA &\hl{60.33}&\hl{76.48}&67.39&\hl{57.01}&\hl{40.42}\\ \hline
        A1&CycleGAN-1G+2DA &\hl{57.64}&\hl{71.49}&64.14&\hl{54.28}&\hl{40.64}\\ \hline
        A1&CycleGAN-1G+3DA &\textbf{\hl{53.29}}&\textbf{\hl{68.61}}&\textbf{\hl{58.46}}&\textbf{\hl{49.48}}&\textbf{\hl{36.61}}\\ \hline \hline
        A2&CycleGAN-2G+2DA & 61.54 & \hl{74.78} & 69.87 &62.14	&\hl{39.39}\\ \hline
        A2&CycleGAN-2G+4DA & \hl{60.88} &\hl{73.73} &67.54 &59.98 &\hl{42.27}\\ \hline
        A2&CycleGAN-2G+6DA & \textbf{\hl{57.53}} &\textbf{\hl{70.48}}	&65.17&\textbf{\hl{54.97}}&\textbf{\hl{39.50}}\\ \hline \hline
        A3&CycleGAN-8G+8DA & \hl{60.62} & \hl{72.87} & 62.42 & 58.71 & 44.79\\ \hline
        A3&CycleGAN-8G+16DA & \hl{55.30} & \hl{66.37} & \hl{60.29} &\hl{51.03} & \hl{37.80} \\ \hline
        A3&CycleGAN-8G+24DA & \textbf{\hl{52.80}} & \textbf{\hl{65.84}} & \textbf{\hl{56.54}} & \textbf{\hl{46.49}} & \textbf{\hl{36.91}}\\ \hline
        A3&CycleGAN-8G+32DA &\hl{54.01}&\hl{67.00}&\hl{56.29}&\hl{49.44}&\hl{36.70} \\ \hline
    \end{tabular}
    \caption{The WERs of baseline ASR w/o and w/ Multi-Discriminators CycleGAN SE on et05\_real\_noisy.}
    \label{tab:tab3}
\end{table}

\subsection{One Generator vs. Multi-Generators}
\begin{figure}[htb]
\begin{minipage}[b]{1.0\linewidth}
  \centering
  \centerline{\includegraphics[width=8.5cm]{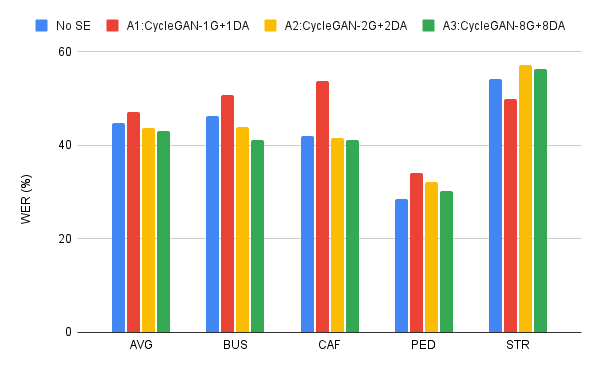}}
  \centerline{(a) The WERs of ASR w/ 1DA models SE}\medskip
\end{minipage}
\begin{minipage}[b]{.48\linewidth}
  \centering
  \centerline{\includegraphics[width=4.0cm]{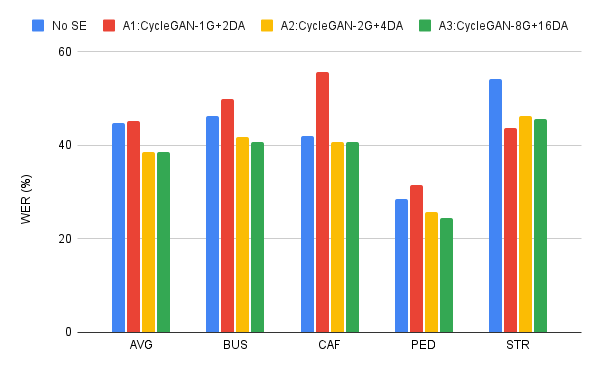}}
  \centerline{(b) w/ 2DA models SE}\medskip
\end{minipage}
\hfill
\begin{minipage}[b]{0.48\linewidth}
  \centering
  \centerline{\includegraphics[width=4.0cm]{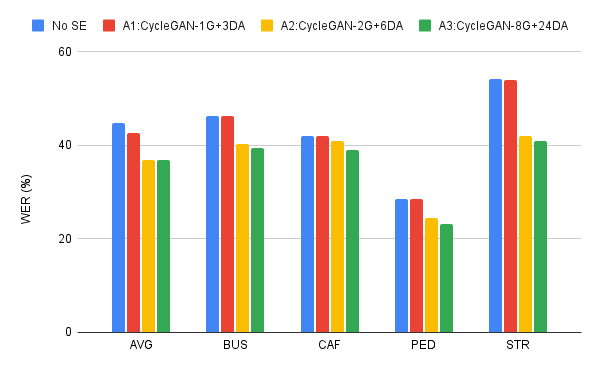}}
  \centerline{(c) w/ 3DA models SE}\medskip
\end{minipage}
\caption{The performance comparisons on the data of female speakers in the development set among the models with one, two and eight generators. 
}
\label{fig:multig}
\end{figure}

Figure \ref{fig:multig} shows the comparisons in terms of WERs between SE systems with different number of generators given the same number of discriminators. For example, Figure \ref{fig:multig} $(a)$ shows the WER comparison between A1:CycleGAN-1G+1DA, A2:CycleGAN-2G+2DA and A3:CycleGAN-8G-+8DA. These models all have only one discriminator for one $G\_A$. Figure \ref{fig:multig} $(b)$ is the comparison between A1:CycleGAN-1G+2DA, A2:CycleGAN-2G+4DA and A3:CycleGAN-8G-+16DA which all have two discriminators for one $G\_A$. The comparison between the models having three discriminators for one $G\_A$, A1:CycleGAN-1G+3DA, A2:CycleGAN-2G+6DA and A3:CycleGAN-8G+24DA, is shown in Figure \ref{fig:multig} $(c)$.

For one discriminator A (1DA) and two discriminators A (2DA) models in Figure \ref{fig:multig}  $(a)$ and Figure \ref{fig:multig} $(b)$, the model with two generators (A2:CycleGAN-2G) performs better than the one with only one generator (A1:CycleGAN-1G) in the BUS, CAF and PED noisy scenarios. The model with eight generators (A3:CycleGAN-8G) slightly performs better than A2:CycleGAN-2G. For three discriminator A (3DA) models in Figure \ref{fig:multig} $(c)$, A3:CycleGAN-8G slightly performs better than A2:CycleGAN-2G, and A2:CyclGAN-2G and A3:CyclGAN-8G models perform much better than A1:CycleGAN-1G. In sum, our results show that adding more generators and train them with well split data, e.g. gender and noise types, plays an important role to improve the overall performance. 

\section{Analysis \& Discussion}
\subsection{WER, Insertion, Deletion and Substitution}
Table \ref{tab:delsubins} shows the comparison (WER, Insertion, Deletion, Substitution and Correction) on CHiME-3 development set with and without Multi-Discriminators CycleGAN SE. 

For all the A1:CycleGAN-1G, A2:CycleGAN-2G and A3:CycleGAN-8G models, the model having only discriminator A (1DA models) has two or three times more insertions than the one having multiple discriminators. It implies that 1DA models do not remove noisy signal very well, so the clean ASR system mis-recognize it as speech. The number of deletion from multiple discriminators models is less than the one from $1DA$ models, which means more words are recognized by the ASR system with Multi-Discriminators CycleGAN SE.
\begin{table}[!htb]
\scriptsize
    \centering
    \begin{tabular}{|c|l|c|c|c|c|c|}
     \hline
    Arch.& Method & WER & INS & DEL & SUB & COR \\\hline
&No SE &  32.14 & \textbf{170} & 6054 & \textbf{2491} & 8715\\\hline
     A1&CycleGAN-1G+1DA &35.25 &723 &\hl{5894} &\hl{2942}&9559\\\hline
     A1&CycleGAN-1G+3DA &\hl{29.45} &240 &\hl{5309} &\hl{2437}&\hl{7986}\\\hline\hline
     A2&CycleGAN-2G+2DA &34.05 & 779 & \hl{5372} & 3083 & 9234\\\hline
     A2&CycleGAN-2G+4DA &\hl{31.44} & 303 & \hl{5611} & 2611 & \hl{8525}\\\hline\hline
     A3&CycleGAN-8G+8DA & 35.12 & 718 & \textbf{\hl{5940}} & 2867 & 9525\\\hline
     A3&CycleGAN-8G+24DA & \textbf{\hl{28.66}} & 398 & \textbf{\hl{4255}} & 3120 & \textbf{\hl{7773}}\\\hline
\end{tabular}
    \caption{The insertion, deletion, substitution and correction on dt05\_real\_noisy set (27119 words) with and without Multi-Discriminators CycleGAN SE.}
    \label{tab:delsubins}
\end{table}

\subsection{ASR outputs \& Spectrograms}
In this section, we compare cherry-picked ASR outputs in the development set without and with Multi-Discriminators CycleGAN SE and the corresponding log-Mel spectrograms. 
Table \ref{tab:ASROut-BUS-del} shows the ASR output on the BUS real noisy utterance without SE, with A1:CycleGAN-1G SE and our best SE model (A3:CycleGAN-8G+24DA). As stated in Table \ref{tab:delsubins}, adding more CycleGAN and discriminator A (DA) for one $G\_A$ helps easing the high deletion problem. 

\begin{figure*}[!htb]
\subfloat[\scriptsize Clean\label{fig:clean-0-500}]
  {\includegraphics[width=.2\linewidth]{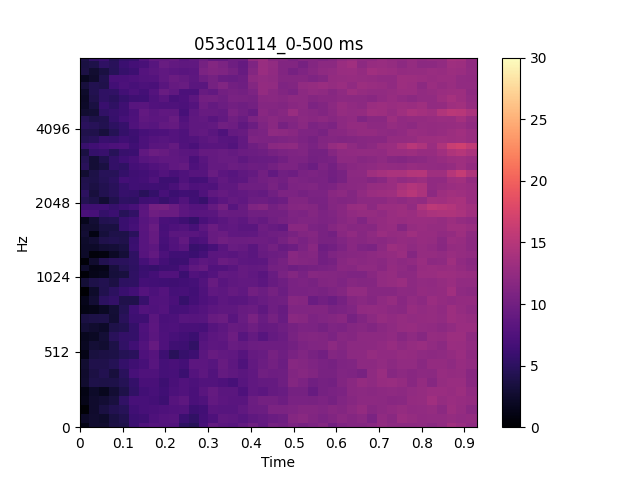}}\hfill
\subfloat[\scriptsize No SE\label{fig:NoSE-0-500}]
  {\includegraphics[width=.2\linewidth]{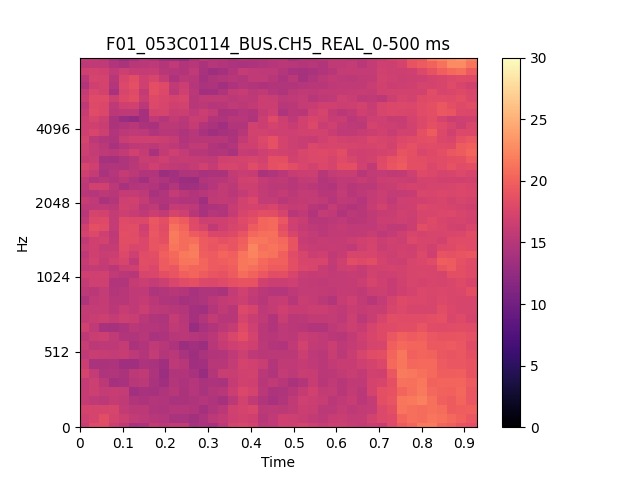}}\hfill
\subfloat[\scriptsize A1:CycleGAN-1G+1DA\label{fig:1G-0-500}]
  {\includegraphics[width=.2\linewidth]{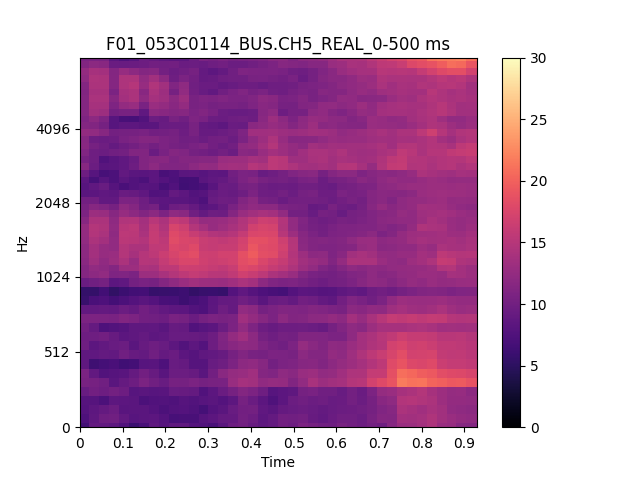}}\hfill
\subfloat[\scriptsize A1:CycleGAN-1G+3DA\label{fig:1G3D-0-500}]
  {\includegraphics[width=.2\linewidth]{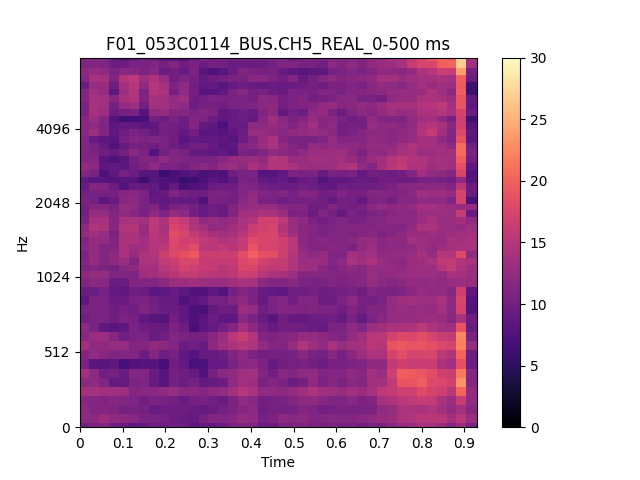}}
\subfloat[\scriptsize A3:CycleGAN-8G+24DA\label{fig:8G24D-0-500}]
  {\includegraphics[width=.2\linewidth]{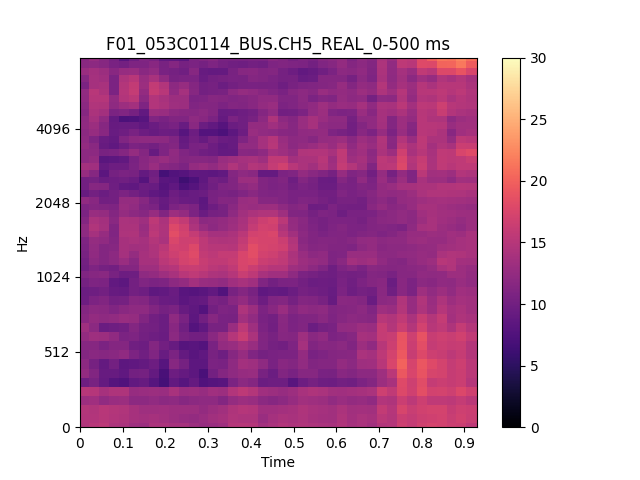}}
\caption{The log-Mel Spectrogram from 0-500 ms from an utterance in BUS development set (the same utterance as in Table \ref{tab:ASROut-BUS-del}) without SE and with Multi-Discriminators CycleGAN SE.}\label{fig:bus-spectrogram-0-500}
\end{figure*}

\begin{figure*}[!htb]
\subfloat[\scriptsize Clean\label{fig:NoSE-0-500}]
  {\includegraphics[width=.2\linewidth]{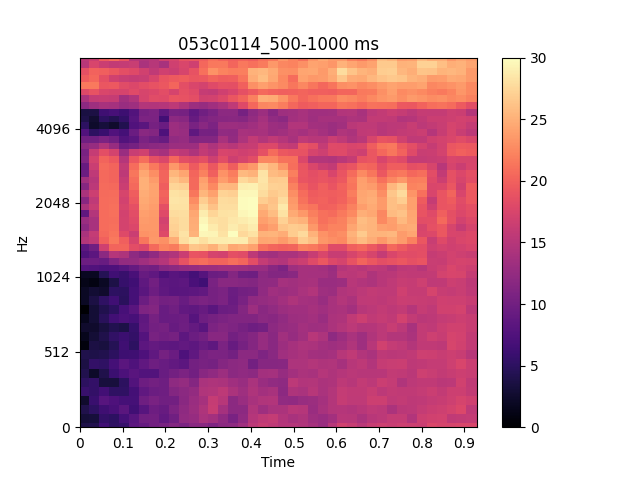}}\hfill
\subfloat[\scriptsize No SE\label{fig:NoSE-500-1000}]
  {\includegraphics[width=.2\linewidth]{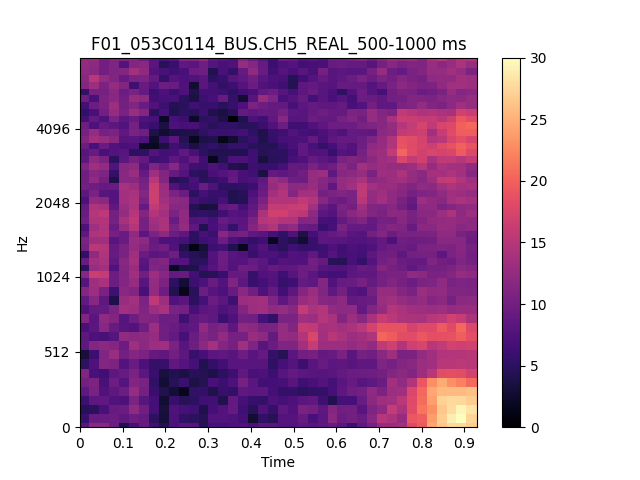}}\hfill
\subfloat[\scriptsize A1:CycleGAN-1G+1DA\label{fig:1G-500-1000}]
  {\includegraphics[width=.2\linewidth]{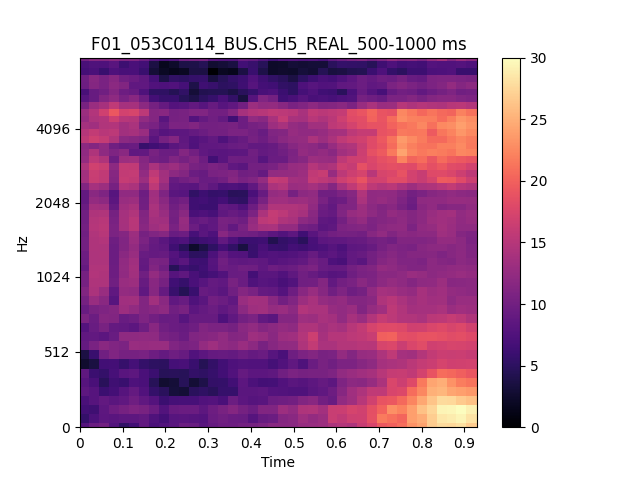}}\hfill
\subfloat[\scriptsize A1:CycleGAN-1G+3DA\label{fig:1G3D-500-1000}]
  {\includegraphics[width=.2\linewidth]{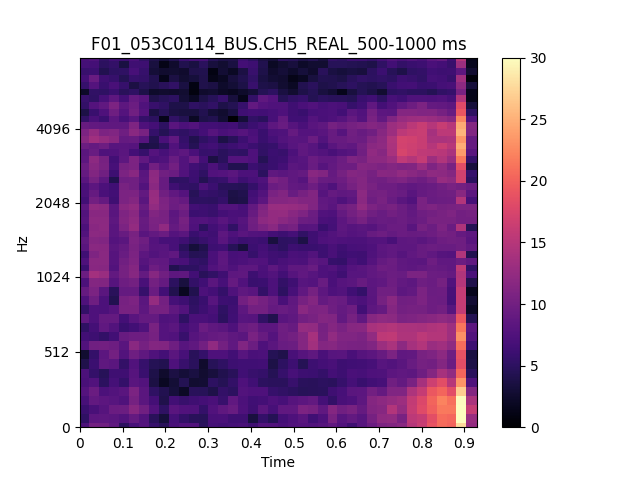}}
\subfloat[\scriptsize A3:CycleGAN-8G+24DA\label{fig:8G24D-500-1000}]
  {\includegraphics[width=.2\linewidth]{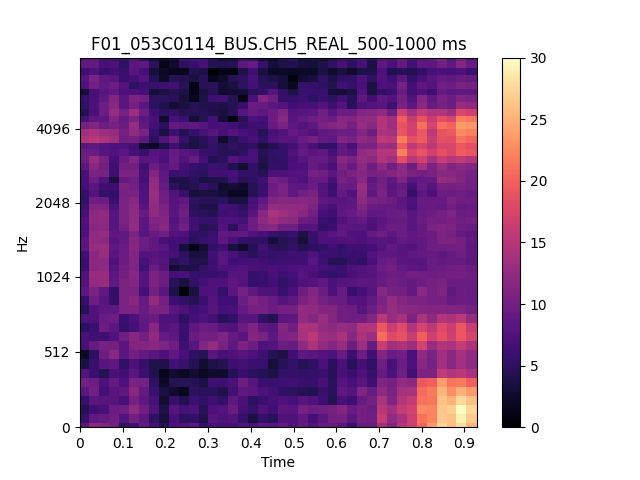}}
\caption{The log-Mel Spectrogram from 500-1000 ms from an utterance in BUS development set (the same utterance as in Table \ref{tab:ASROut-BUS-del}) without SE and with Multi-Discriminators CycleGAN SE.}\label{fig:bus-spectrogram-500-1000}
\end{figure*}

\begin{figure*}[!htb]
\subfloat[\scriptsize Clean\label{fig:NoSE-0-500}]
  {\includegraphics[width=.2\linewidth]{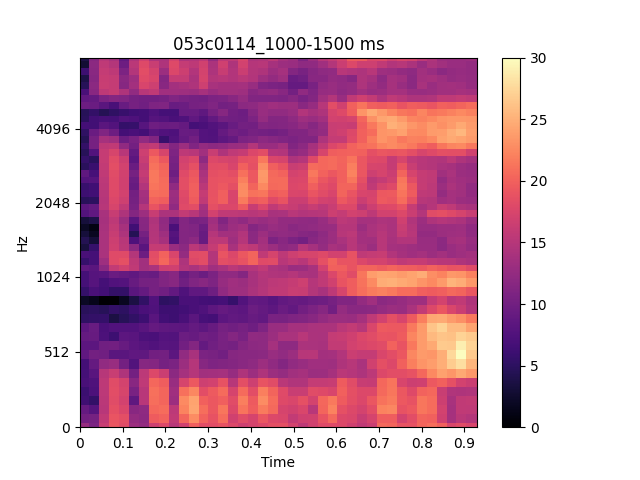}}\hfill
\subfloat[\scriptsize No SE\label{fig:NoSE-1000-1500}]
  {\includegraphics[width=.2\linewidth]{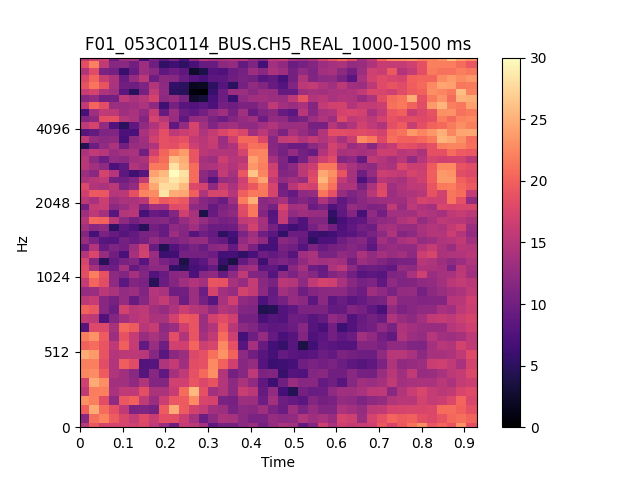}}\hfill
\subfloat[\scriptsize A1:CycleGAN-1G+1DA\label{fig:1G-1000-1500}]
  {\includegraphics[width=.2\linewidth]{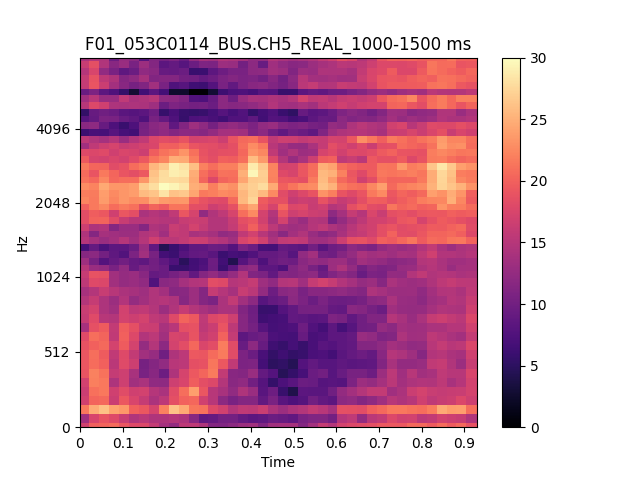}}\hfill
\subfloat[\scriptsize A1:CycleGAN-1G+3DA\label{fig:1G3D-1000-1500}]
  {\includegraphics[width=.2\linewidth]{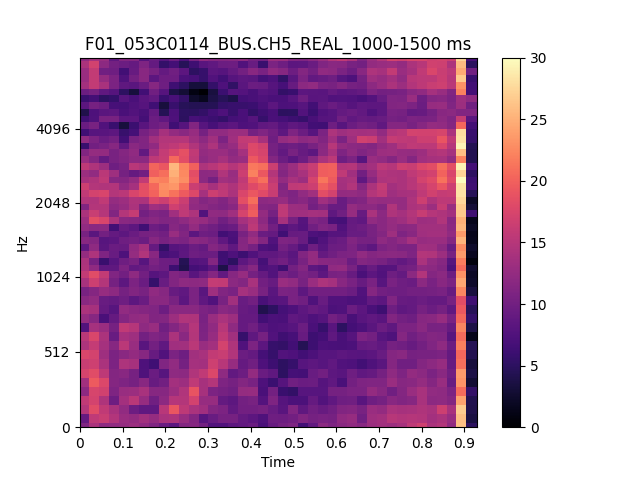}}
\subfloat[\scriptsize A3:CycleGAN-8G+24DA\label{fig:8G24D-1000-1500}]
  {\includegraphics[width=.2\linewidth]{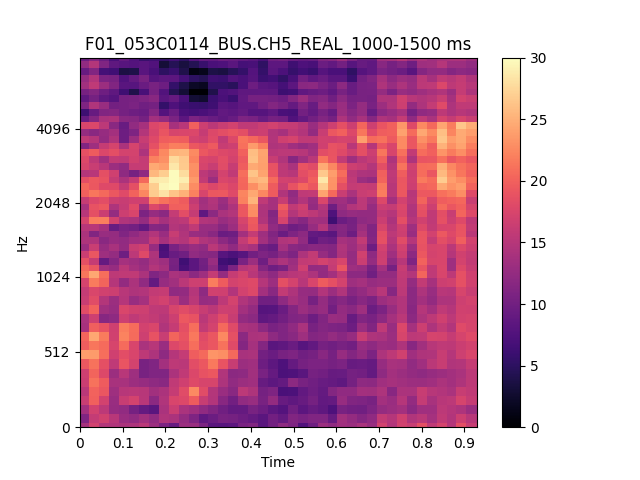}}
\caption{The log-Mel Spectrogram from 1000-1500 ms from an utterance in  BUS development set (the same utterance as in Table \ref{tab:ASROut-BUS-del}) without SE and with Multi-Discriminators CycleGAN SE.}\label{fig:bus-spectrogram-1000-1500}
\end{figure*}

Figure \ref{fig:bus-spectrogram-0-500}, \ref{fig:bus-spectrogram-500-1000} and  \ref{fig:bus-spectrogram-1000-1500} show the comparison of log-Mel Spectrogram from orignal clean utterance, noisy utterance without SE, noisy utterance with A1:CycleGAN-1G+1DA, noisy utterance with A1:CycleGAN-1G+3DA and noisy utterance with A3:CycleGAN-8G+24DA SE. Between 0 to 1000 millisecond, our SE models reduce the noise and help the ASR to recognize the words successfully. However, from 1000 to 1500 milliseconds, the utterance with Multi-Discriminators CycleGAN SE might still contain some noisy signals, so the ASR could not recognize the two words "WANT THEM" well. Besides, A1:CycleGAN-1G+1DA might not transform the noisy signal to clean signal well, so that the ASR system mis-recognizes /A/ as /O/, and the latter predictions ("GIVE UP") have nothing in common with the two words "WANT THEM" with respect to the pronunciation. 

Table \ref{tab:ASROut-CAF} shows another example of ASR output in CAF (SNR=10) noisy development set with and without Multi-Discriminators CycleGAN SE. The A3:CycleGAN-8G model performs much better than the A1:CycleGAN-1G models, there are many substitutions and insertions in A1:CycleGAN-1G models. But the A3:CycleGAN-8G model has more deletions than A1:CycleGAN-1G models in this example.

\begin{table}[!htb]
    \centering
    \scriptsize
    \begin{tabular}{|c|l|l|}
     \hline
     Arch.& Method  & ASR output \\\hline
     &Time            & 0-1000    ms: OUR CUSTOMERS  \\
     &                & 1000-1500 ms: WANT THEM \\\hline
     &Reference       &  OUR CUSTOMERS WANT THEM \\\hline
     &No SE          &  \\\hline
     A1&CycleGAN-1G+1DA &  OUR CUSTOMERS \hl{WON'T GIVE UP}\\\hline
     A1&CycleGAN-1G+3DA &  OUR CUSTOMERS \\\hline
     A3&CycleGAN-8G+24DA & OUR CUSTOMERS\\\hline
    \end{tabular}
    \caption{ASR outputs of an example in the BUS real noisy (SNR=7) development set with and without Multi-Discriminators CycleGAN SE.}
    \label{tab:ASROut-BUS-del}
\end{table}

\begin{table}[!htb]
    \scriptsize
    \centering
    \begin{tabular}{|c|l|l|}
     \hline
     Arch.& Method  & ASR output \\\hline
     &Reference       &  WHAT ABOUT IN SOUTH AFRICA \\
     &                &  ITSELF\\\hline
     &No SE.          &  WHAT ABOUT \hl{A SATISFACTORY}\\\hline
     A1&CycleGAN-1G+1DA &  \hl{BUT} WHAT ABOUT \hl{A} SOUTH \\
      &               & \hl{AFRICAN} \hl{GETS OUT}\\\hline
     A1&CycleGAN-1G+3DA &  WHAT ABOUT \hl{A SAD FACT GETS OUT} \\\hline
     A3&CycleGAN-8G+24DA & WHAT ABOUT SOUTH AFRICANS \\\hline
    \end{tabular}
    \caption{ASR outputs of an example in the CAF (SNR=10) noisy development set with and without Multi-Discriminators CycleGAN SE.}
    \label{tab:ASROut-CAF}
\end{table}

\section{Conclusion}
In this work, we investigate the performance of ASR in terms of WER using CycleGAN based feature mapping and our novel extension (multiple generators and multiple discriminators). Our experimental results show that multiple generators trained with well splitting subset is better than one generator trained with all the data. 
Besides, the models have multiple discriminator A ($D\_A$) per generator A ($G\_A$) improve the average WER. The best model A3:CycleGAN-8G+24DA improves the WER for all the four noisy scenarios and achieves 10.03 \% relatively WER on development set and 14.09 \% relatively WER on evaluation set.

\pagebreak

\bibliographystyle{IEEEbib}
\bibliography{strings,refs}

\end{document}